\documentclass[conference]{IEEEtran}
\IEEEoverridecommandlockouts

\usepackage{cite}
\usepackage{amsmath,amssymb,amsfonts}
\usepackage{algorithmic}
\usepackage{graphicx}
\usepackage{textcomp}
\usepackage{xcolor}
\usepackage{booktabs}

\def\BibTeX{{\rm B\kern-.05em{\sc i\kern-.025em b}\kern-.08em
    T\kern-.1667em\lower.7ex\hbox{E}\kern-.125emX}}
\begin{document}
\title{Modeling Cultural Bias in Facial Expression Recognition with Adaptive Agents\\
}

\author{\IEEEauthorblockN{David Freire-Obregón}
\IEEEauthorblockA{\textit{SIANI, ULPGC} \\
Spain \\
0000-0003-2378-4277}
\\
\IEEEauthorblockN{Oliverio J. Santana}
\IEEEauthorblockA{\textit{SIANI, ULPGC} \\
Spain \\
0000-0001-7511-5783}
\and
\IEEEauthorblockN{José Salas-Cáceres}
\IEEEauthorblockA{\textit{SIANI, ULPGC} \\
Spain \\
0009-0004-7543-3385}
\\
\IEEEauthorblockN{Daniel Hernández-Sosa}
\IEEEauthorblockA{\textit{SIANI, ULPGC} \\
Spain \\
0000-0003-3022-7698}
\and
\IEEEauthorblockN{Javier Lorenzo-Navarro}
\IEEEauthorblockA{\textit{SIANI, ULPGC} \\
Spain \\
0000-0002-2834-2067}
\\
\IEEEauthorblockN{Modesto Castrillón-Santana}
\IEEEauthorblockA{\textit{SIANI, ULPGC} \\
Spain \\
0000-0002-8673-2725}
}

\maketitle

\begin{abstract}
Facial expression recognition (FER) must remain robust under both cultural variation and perceptually degraded visual conditions, yet most existing evaluations assume homogeneous data and high-quality imagery. We introduce an agent-based, streaming benchmark that reveals how cross-cultural composition and progressive blurring interact to shape face recognition robustness. Each agent operates in a frozen CLIP feature space with a lightweight residual adapter trained online at $\sigma=0$ and fixed during testing. Agents move and interact on a $5\times5$ lattice, while the environment provides inputs with $\sigma$-scheduled Gaussian blur. We examine monocultural populations (Western-only, Asian-only) and mixed environments with balanced (5/5) and imbalanced (8/2, 2/8) compositions, as well as different spatial contact structures. Results show clear asymmetric degradation curves between cultural groups: JAFFE (Asian) populations maintain higher performance at low blur but exhibit sharper drops at intermediate stages, whereas KDEF (Western) populations degrade more uniformly. Mixed populations exhibit intermediate patterns, with balanced mixtures mitigating early degradation, but imbalanced settings amplify majority-group weaknesses under high blur. These findings quantify how cultural composition and interaction structure influence the robustness of FER as perceptual conditions deteriorate.

\end{abstract}


\section{Introduction}

Facial expressions are a significant channel for communicating affect, intent, and social information \cite{Ekman71}. Rapid interpretation of facial cues enables trust, coordination, and affect-aware interaction. For AI systems running in human environments, dependable facial expression recognition (FER) is necessary for socially aware behavior.

\begin{figure}[ht]
    \centering
    \includegraphics[width=0.6\columnwidth]{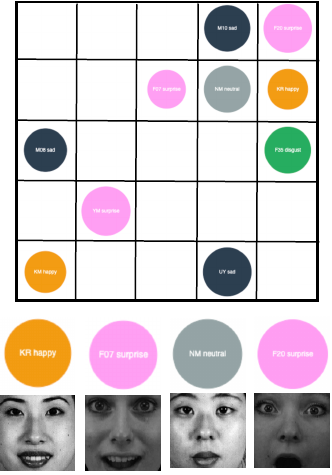}
    \caption{Initial snapshot (t = 1) of the agent lattice. Agents occupy grid cells as colored circles: color indicates the current emotion (e.g., happy, sad, angry), and the text inside each circle denotes the agent’s ID and the elicited emotion.}
    \label{fig:intro1}
\end{figure}

\begin{figure}[ht]
    \centering
    \includegraphics[width=0.6\columnwidth]{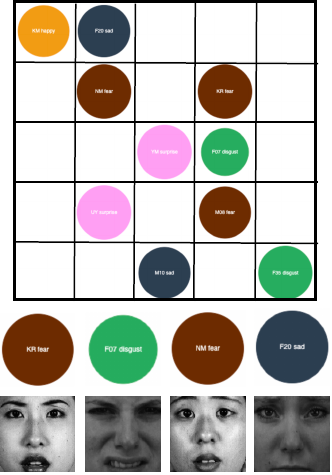}
    \caption{Agent lattice at t = 5. Circle size reflects confidence in their emotion predictions: repeated misclassifications shrink the circle, while each correct prediction expands it back toward its original size.}
    \label{fig:intro2}
\end{figure}

Recent advances in computer vision, from deep CNNs to vision-language encoders like CLIP, have improved FER accuracy on conventional benchmarks \cite{Mollahosseini16}. Most, however, do so under the assumption of clean imagery and culturally homogeneous data; a scenario rarely encountered in the real world. Two factors are particularly troublesome: (i) cross-cultural variation in facial structure, display rules, and learned perceptual biases \cite{Elfenbein02}, and (ii) perceptual degradation (e.g., blur) that weakens subtle cues \cite{Dhall15}. Models trained on a single population (e.g., Western faces) may tend to perform poorly on a different one (e.g., Japanese faces), and quality loss can exacerbate the issue.

To explain both of these effects simultaneously, a protocol must be developed that transcends simple, static train-once/test-once evaluations. We propose a streaming, agent-focused method: agents process a sequence of inputs with varying quality over time, and we make judgments about performance as the conditions change. The perspective is proportional to real-world deployments and reveals gaps in groups under degradation.

We present an agent-based model of cross-cultural facial expression perception in the presence of $\sigma$-scheduled Gaussian blur. Agents each have a randomly chosen unique identity taken from one of two culturally distinct datasets (see Figure \ref{fig:intro1}). Every fixed number of ticks, the agent outputs a facial expression taken randomly from the repertoire of that personality so that agents nearby can perceive and classify them. Perception entails frozen CLIP embeddings \cite{radford21}, then a lightweight residual adapter that is only trained during a preliminary learning period at $\sigma=0$, and finally frozen during testing. Agents live on a $5\times5$ lattice and see a stream cycling through higher $\sigma$ levels.

We create experimental setups with monocultural populations (Western-only, Asian-only) and mixed populations (balanced $5/5$ and imbalanced $8/2$, $2/8$) as seen in Figures \ref{fig:intro1} and \ref{fig:intro2}. We observe asymmetric robustness curves between Western and Asian groups across settings, with $\sigma$-dependent performance gaps that widen as blur increases. We find that mixed populations and contact structure influence both accuracy and calibration, and demonstrate where cross-cultural exposure is beneficial and where it is not. Our contributions are summarized as follows:

\begin{itemize}
\item We model multicultural settings with agents that each represent a unique person identity drawn from different cultural groups, and that actively display random expressions over time on a $5\times5$ lattice, enabling neighbors to perceive and classify them. This captures the dynamics of facial expression exchange in culturally mixed environments.
\item We analyze how cross-cultural mixture (monocultural vs.\ mixed) define recognition within $\sigma$-scheduled blur, reporting functionality metrics.
\item All agents employ frozen CLIP embeddings with a lightweight residual adapter trained only at $\sigma{=}0$ and kept frozen during testing. In this context, adaptivity does not refer to online parameter updates but to the agents’ dynamic behavioral adaptation, through interaction, confidence exchange, and exposure to changing perceptual conditions, rather than to model fine-tuning. This design reflects realistic bootstrapping followed by deployment in progressively more complex environments.
\item We observe the asymmetric breakdown of various cultural groups, $\sigma$-dependent gaps that increase with blur, and situations involving population mixture that either reduce or widen these gaps. These findings can be applied to the design of human–machine interaction and affect-aware systems across different settings.
\end{itemize}

\section{Related Work}

We set our study at the intersection of three strands. First, research on facial expressions spans psychological theories and computational FER, and is central to understanding how cultural variation in production and perception shapes recognition performance. Second, agent-based approaches provide a natural lens for analyzing perception and interaction as local, temporally evolving processes on a spatial substrate, which is particularly well-suited for studying contact structure and information exchange. Third, the literature on bias in FER documents systematic performance and calibration gaps across demographic and cultural groups, raising questions about how these gaps evolve under changing visual conditions. Our work connects these threads by analyzing multicultural FER with agents that display and perceive expressions. At the same time, image quality degrades via $\sigma$-scheduled blur, and by reporting block-wise metrics that make robustness and group disparities explicit.

\subsection{Facial Expressions}
Classical accounts argue for a small set of basic emotions with characteristic facial configurations that are widely recognizable across cultures \cite{ekman1992basic}. Subsequent cross-cultural perception studies, however, have documented systematic differences between cultural groups in both the production and perception of facial expressions, for example, distinct gaze strategies (eyes/mouth vs.\ center-of-face) and culturally tuned internal representations \cite{blais2008plos,jack12,Salas-Caceres2024}. In computer vision, large-scale datasets such as AffectNet and RAF-DB have expanded FER beyond lab-controlled, posed imagery to \textit{in-the-wild} settings with greater demographic diversity. However, they also expose issues of labeling subjectivity and demographic imbalance that complicate cross-cultural generalization \cite{affectnet2017,rafdb2017}. Nevertheless, these datasets do not contain repeated instances of the same subjects systematically portraying the complete set of seven basic facial expressions, which limits their suitability for agent-based simulations that require consistent identity–expression mappings. To endow each agent with a stable and unique identity across all expression categories, it is preferable to employ culturally distinct yet structurally comparable datasets such as KDEF and JAFFE \cite{kdef1998,jaffe1998}, where each subject deliberately enacts the complete set of basic emotions under controlled acquisition conditions. This design facilitates cross-cultural comparisons while preserving consistency of identity, making these datasets well-suited for modeling individual agents with fixed expressive repertoires.

\subsection{Agents}

Agent-based models (ABM) offers a principled approach to studying emergent social phenomena through local interaction rules among autonomous entities \cite{Epstein96, Bonabeau02}. Rather than starting from overarching system rules, ABMs adopt a bottom-up perspective: they define simple behaviors for individual agents, and the overall system dynamics naturally emerge from their interactions. 
For perception-centric problems, ABM allows us to endow agents with heterogeneous perceptual modules, spatial neighborhoods, and interaction dynamics, and then examine how system-level properties (accuracy, calibration, group disparities) evolve as environmental conditions change (e.g., increasing blur). They have been effectively utilized across diverse fields such as economics \cite{hamill16}, political science \cite{Moya17}, and trust-oriented social networks \cite{Chica18}, among others.

Recent studies have shown how recognition degradation and recognition bias can alter collective behavior: the misinterpretation of affective cues leads to trust collapse and polarization in artificial societies \cite{freire25wrong}, and a step-by-step loss of perceptual accuracy results in symbolic–perceptual dissociation and social instability \cite{freire25fading}. This line of research is particularly relevant to multicultural FER because the recognition outcome is determined by identity, contact structure, and local observation.

\subsection{Bias in FER}
In the last decade, FER has experienced rapid progress thanks to advances in deep learning techniques and the availability of larger and more diverse datasets \cite{Onyema21}, which have expanded its applicability in real-world contexts \cite{Dhall15}. For instance, FER technologies are now employed in various domains, including monitoring the emotional and physical states of industrial workers \cite{Patwardhan16}.

However, this unprecedented expansion has been accompanied by growing concerns about fairness and bias. These concerns are not specific to FER but have long been documented for face analysis more broadly, with seminal works showing dramatic accuracy differentials across demographic categories in commercial systems \cite{Buolamwini18}. In FER in particular, current studies highlight ethnicity- and group-based error behaviors, including asymmetric confusions and persistent performance differentials, even when using modern encoders, as well as sensitivity to dataset composition and balance \cite{Chhua24}.

Nevertheless, even with their increasing impact, FER systems remain troubled with a significant challenge: widespread cultural bias \cite{FlejedeFan23}. These biases frequently remain hidden behind aggregated metrics, which hinders advances towards just and resilient systems. Our contribution is to further address this gap by presenting an agent-based model that simulates cross-cultural populations under different perceptual contexts, allowing for a dynamic examination of the formation and progression of bias in real-time.


\section{Methodology}

\subsection{Overview}
We consider a population of $N$ agents $\mathcal{A}=\{a_1,\dots,a_N\}$ placed on a two–dimensional toroidal lattice
$\mathcal{G}=\{0,\dots,W\!-\!1\}\times\{0,\dots,H\!-\!1\}\subset\mathbb{Z}^2$.
Time is discrete, $t\in\mathbb{N}$. Each agent occupies a single cell (hard exclusion) and interacts locally with its Moore
neighborhood $\mathcal{N}_i(t)\subset\mathcal{A}$ (the eight adjacent cells). At every simulation tick, each agent \emph{displays} a facial
expression tied to its identity; neighbors perceive and classify that expression under a controlled level of Gaussian blur.

The agent-based formulation was deliberately chosen over static population-level analyses to capture emergent effects arising from unequal exposure, local learning, and structured contact. Unlike conventional batch experiments, ABM allows us to observe how micro-level interactions (i.e., confidence propagation, peer learning, and neighborhood composition) shape the evolution of recognition bias dynamically over time.

\subsection{Cultural identities and expression display}
Let $\mathcal{E}=\{\text{neutral},\text{happy},\text{sad},\text{anger},\text{disgust},\text{fear},\text{surprise}\}$ be the label set.
Each agent $a_i$ is assigned:
\[
a_i \;=\; \langle \text{group } g_i,\;\text{identity } \mathrm{id}_i,\;\mathcal{E}_i,\;\vec p_i(t) \rangle,
\]
Where $g_i\in\{\text{Western},\text{Asian}\}$ is the cultural group, $\mathrm{id}_i$ is a unique person identity sampled from that
group, and $\mathcal{E}_i\subseteq\mathcal{E}$ is the set of available expressions for that identity. At each tick, the agent chooses a displayed face $d_i(t)\in\mathcal{E}_i$ uniformly from the set of available facial expressions from the assigned identity. This design yields multicultural populations
(monocultural or mixed compositions) where identities repeatedly express varying emotions over time, enabling local perception.

\begin{figure}[ht]
    \centering
    \includegraphics[width=0.8\columnwidth]{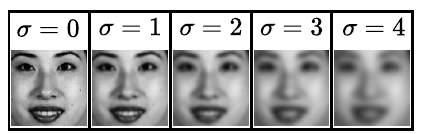}
    \caption{\textbf{Gaussian blur levels (\(\sigma\)).} The same face shown with increasing blur from left to right: \(\sigma = 0,1,2,3,4\). As \(\sigma\) increases, edges soften and fine facial details disappear, making emotion classification more difficult.}
    \label{fig:blur}
\end{figure}

\subsection{Degradation schedule (blur)}
The environment provides blurred inputs through a piecewise–constant schedule $\sigma(t)\in\Sigma$, where $\Sigma=\{0,1,2,3,4\}$, see Figure \ref{fig:blur}.
There is an initial \emph{learning phase} of length $T_{\text{learn}}$ at $\sigma(t)=0$, followed by evaluation blocks with
increasing blur:

\[
\sigma(t) \;=\;
\begin{cases}
0, & 0 \le t < T_{\text{learn}},\\[1ex]
\sigma_b, & t \in [\tau_b, \tau_{b+1}),\quad 
             b = 1,\dots,B,\\
         & \sigma_1 < \sigma_2 < \cdots < \sigma_B.
\end{cases}
\]

For an agent $a_j$ displaying $d_j(t)$ at blur $\sigma(t)$, the environment renders an image $x_j(t)$ by applying isotropic
Gaussian blur of standard deviation $\sigma(t)$ to the identity's base image for $d_j(t)$.

\subsection{Perception model}
Perception runs on frozen embeddings within a fixed feature space.
Each agent \(a_i\) possesses an independent two-layer multilayer perceptron (MLP) classifier \(f_i:\mathbb{R}^D\to\mathbb{R}^{|\mathcal{E}|}\), consisting of layer normalization, a hidden linear transformation with nonlinearity, dropout, and a final linear layer followed by a softmax.
Given a neighbor’s displayed embedding \(x_j^{\text{self}}(t)\), the agent outputs
\[
p_i(x_j^{\text{self}}) = \mathrm{Softmax}\big(f_i(x_j^{\text{self}})\big),
\qquad
\hat e^{(i)}_j(t) = \arg\max_{c\in\mathcal{E}} [p_i]_c,
\]
and records both the predicted emotion and its confidence score.

\paragraph{Learning and test.}
Only the adapter $A_i$ is trained, and only during the learning phase at $\sigma=0$ using cross–entropy:
\[
\min_{W_i}\; \mathbb{E}_{(x,y)\sim \mathcal{D}_{\sigma=0}}\left[-\log p_{i}(x)_y\right].
\]
After $t\ge T_{\text{learn}}$, all $A_i$ are frozen; evaluation proceeds while $\sigma(t)$ increases across blocks.

\subsection{Learning protocol}
Training is carried out online and autonomously for all agents.
In the learning phase (\(\sigma=0\)), each robot optimizes its own detector using its own samples, with the cross-entropy loss between the predicted distribution and the ground-truth label minimized. Additionally, a peer-learning mechanism is enabled: if an agent classifies a neighbor's display with high confidence, it uses that neighbor's sample and label as an extra training example. Following this training stage, the classifiers are frozen. At each time step, an agent classifies either all of its neighbors, given that such neighbors have yielded valid displays in the current time step.

\subsection{Interaction and movement}
At each simulation tick, all agents perform the following steps:
\begin{enumerate}
    \item It randomly displays an emotional embedding drawn from its identity stream at its blur level.
    \item Each agent classifies the item, all neighbors, and a peer-learning update is performed if the confidence criterion is met.
    \item Agents measure the valence of emotions in their local neighborhood. Let 
    \(\text{NEG}=\{\text{anger},\text{disgust},\text{fear},\text{sad}\}\) and 
    \(\text{POS}=\{\text{happy},\text{surprise},\text{neutral}\}\).
If the number of negative agents exceeds the number of positive agents by a constant threshold value, then the agent attempts to migrate towards an empty adjacent cell (avoiding); otherwise, it migrates blindly with a constant probability. Moves succeed only if the destination cell is empty.
\end{enumerate}

\section{Experimental Setup}

This section describes the experimental design used to explore FER in the context of cultural changes and visual degradation. We first introduce the training and test datasets for the agents’ emotion classifiers. We then present the agent-based simulation on a toroidal lattice of size \(5\times 5 \), and provide a brief description of the perception–learning pipeline. 

\textbf{Datasets.} We trained our agents' classifiers using two standard facial‐expression datasets: JAFFE and KDEF. The Japanese Female Facial Expression (JAFFE) dataset \cite{jaffe1998} comprises facial images of 10 Japanese female subjects, each depicting the six basic emotions (anger, disgust, fear, happiness, sadness, and surprise) as well as neutral expressions. Due to its cultural and demographic homogeneity, JAFFE is a valuable tool for examining cross-population generalization. The Karolinska Directed Emotional Faces (KDEF) database \cite{lundqvist1998kdef} consists of high-resolution, frontal portraits of 70 stage actors, collected under strict control of conditions, each endorsing the same seven classes. From each sequence, we considered two frames (neutral and apex expression), and thus our balanced set consists of six times the six non-neutral emotions, plus the neutrals. We acknowledge that JAFFE and KDEF differ not only in cultural provenance but also in imaging characteristics, such as resolution, lighting, and expression elicitation protocols. Consequently, these datasets serve as operational proxies for culturally distinct facial distributions rather than perfect sociocultural representations. The study thus focuses on how differences in distributional structure, whether cultural or stylistic, affect robustness in recognition.

\textbf{Implementation Details.}
We consider a \(5\times5\) toroidal lattice with hard exclusion (at most one per cell) and Moore neighborhoods up to eight wrap-around neighbors. Considering the number of identities in both datasets, we study four composition populations with \(N=10\) agents: \emph{Western-only} (10/0), \emph{Asian-only} (0/10), and \emph{mixed} with balanced (5/5) and imbalanced (8/2, 2/8) identity allocation. Each agent receives a unique person identity from its assigned cultural group. At tick \(t\), the agent \emph{displays} one labeled sample \((x^{\text{self}},y^{\text{self}})\) chosen uniformly over its identity-specific stream at the currently active blur \(\sigma(t)\); sampling over the stream is uniformly random and constrained by the current \(\sigma\) level.

Time is discrete in our simulations. We run a learning phase of \(T_{\text{learn}}=1000\) ticks at \(\sigma=0\), followed by evaluation blocks of fixed length \(T_{\text{block}}=200\) ticks that sweep \(\sigma \in \{0,1,2,3,4\}\) in increasing order. Block boundaries occur at \(t=T_{\text{learn}}\) and at every change of \(\sigma\); block-level metrics are reset at those boundaries. 

Perception operates on frozen \(D\)-dimensional visual embeddings. Each agent owns an independent two-layer MLP classifier \(f_i:\mathbb{R}^D\!\to\!\mathbb{R}^{7}\) with the following stack: LayerNorm \(\rightarrow\) Linear(\(D\!\to\!H\)) \(\rightarrow\) GELU \(\rightarrow\) Dropout \((p=0.1)\) \(\rightarrow\) Linear(\(H\!\to\!7\)) \(\rightarrow\) Softmax, where the hidden size is \(H=512\). Given a neighbor’s display \(x_j^{\text{self}}(t)\), the agent outputs \(p_i=\mathrm{Softmax}(f_i(x_j^{\text{self}}))\), predicts \(\hat e^{(i)}_j(t)=\arg\max_c [p_i]_c\), and records the maximum probability as confidence.

Training is online and per-agent. During learning (\(\sigma=0\)), each agent updates its own MLP, minimizing cross-entropy with label smoothing \(0.05\). Optimization uses AdamW with learning rate \(3\times 10^{-4}\), weight decay \(5\times 10^{-2}\). After learning, classifiers are frozen for evaluation by default. At every tick, an agent categorises all currently free neighbors in its Moore neighborhood. A smoothed correctness trace (“trust”) is maintained for visualization purposes only (as seen in Figure \ref{fig:intro2}).

If \(\#\text{NEG}-\#\text{POS}\ge 2\) in the neighborhood, then the agent tries moving to an adjacent free cell (avoiding). Otherwise, it makes a random step with probability \(0.7\). Only moves can proceed into vacant cells. 

\section{Experimental Evaluation}

In this section, we analyze how cultural composition and image degradation jointly influence recognition. As previously stated, we compare four cohort arrangements on the same $5{\times}5$ toroidal grid: (i) \emph{single-culture} groups (KDEF-only, JAFFE-only), (ii) the \emph{balanced mixed} collection (5/5), and (iii–iv) two \emph{imbalanced mixed} populations (KDEF-majority 8/2 and JAFFE-majority 2/8). Red vertical lines in the plots indicate the end of the learning phase and the beginning of evaluation at $\sigma{=}0$, an easier regime where performance improves before degradation becomes an issue. Purple vertical lines each subsequent increase in blur level ($\sigma: 0 \!\rightarrow\! 1 \!\rightarrow\! 2 \!\rightarrow\! 3 \!\rightarrow\! 4$). Observe how the first test block at $\sigma{=}0$ commences exactly at the red line; hence, there is no additional purple indicator for this transition, as the red line coincides with the $\sigma{=}0$ onset.

\begin{figure}[ht]
    \centering
    \includegraphics[width=1\columnwidth]{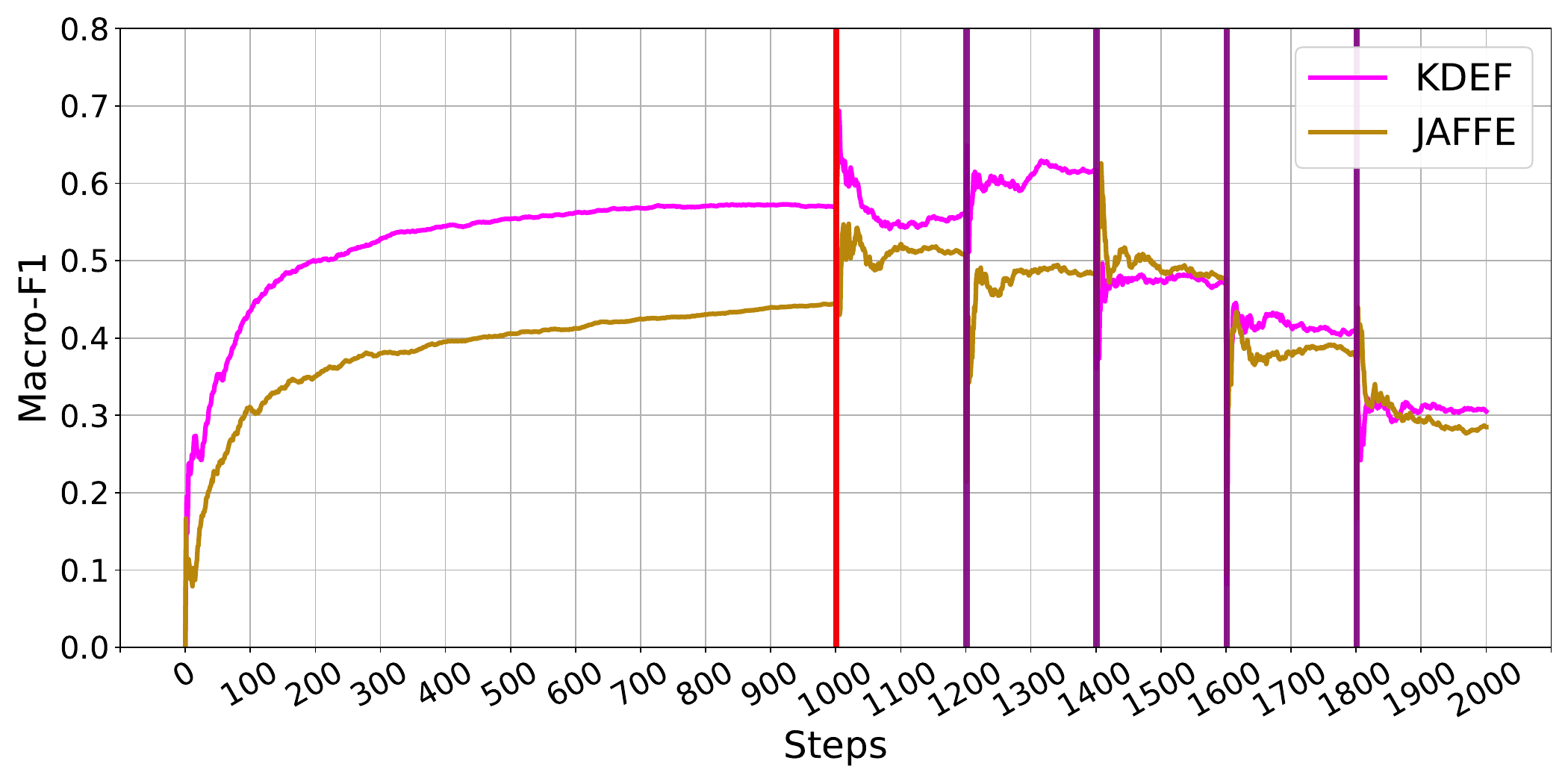}
    \caption{Macro-F1 evolution for monocultural populations (KDEF-only vs.\ JAFFE-only) during learning and evaluation with increasing blur $\sigma \in \{0,1,2,3,4\}$. Red vertical line marks the end of the learning phase.}
    \label{fig:global_pure}
\end{figure}

\subsection{Single-Culture Cohorts: KDEF-only and JAFFE-only}

We begin analyzing monocultural groups consisting entirely of KDEF (Western) or JAFFE (Asian) faces. Figure~\ref{fig:global_pure} demonstrates the evolution of block-wise Macro-F1 over the training period and then the test period with progressively higher blur levels. In the early learning stage ($t < 1000$, $\sigma = 0$), both groups demonstrate an abrupt improvement as agents learn to map their residual adapters from clean, high-quality faces to their respective cultural faces. The performance converges towards the end of the training stage, with an overall Macro-F1 of around $0.70$ for KDEF and $0.58$ for JAFFE, registering slightly faster convergence and subsequent peaks for Western faces.

When evaluation begins ($t \geq 1000$) at $\sigma = 0$, both groups maintain training performance, but differences grow deeper as blur increases. As $\sigma$ increases from $1$ to $4$, performance on recognition degrades slowly for both groups, but the degradation curves are not balanced. KDEF-only populations exhibit a gradual decrease, with acceptable performance lasting until $\sigma = 2$ and a noticeable decline at higher blur intensities. On the other hand, JAFFE-only agents demonstrate an earlier and abrupt decrease, with Macro-F1 descending sharply after $\sigma = 1$ and achieving diminished asymptotic measures by $\sigma = 4$. This implies that JAFFE expression recognition is highly vulnerable to the degradation of high-frequency facial detail and might be due to stylistic variations in expressive behaviors across cultures or dataset properties (such as reduced contrast in JAFFE).

In summary, these monocultural baselines reveal both the upper bound of agent performance in optimal circumstances and the inherent variation in robustness across cultures under controlled perceptual deterioration.
\begin{figure}[ht]
    \centering
    \includegraphics[width=1\columnwidth]{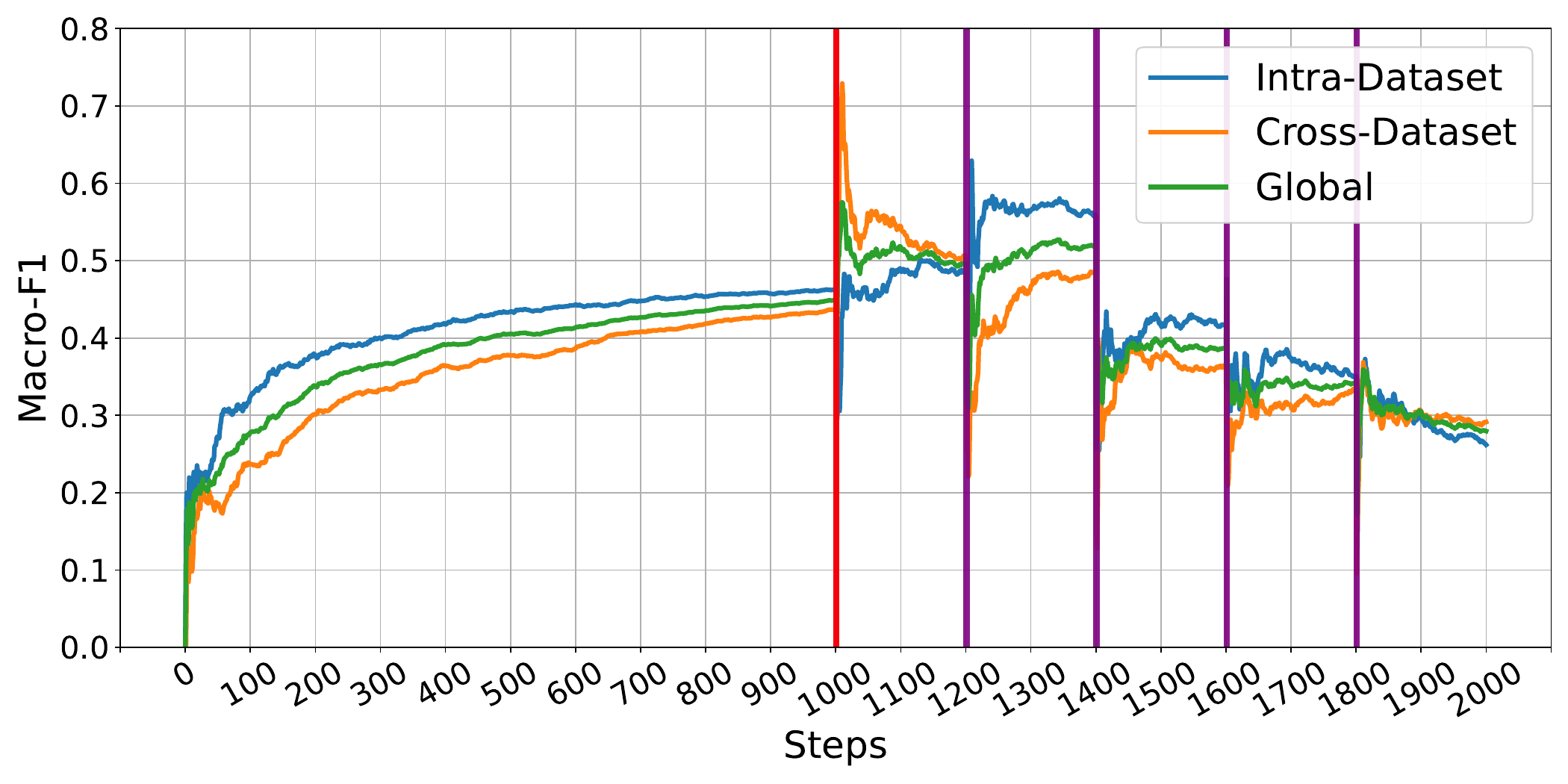}
    \caption{Macro-F1 evolution for a balanced mixed population (5 KDEF / 5 JAFFE). Curves show intra-cultural (same group), cross-cultural (different groups), and global averages as blur $\sigma$ increases.}
    \label{fig:mixed_5_5}
\end{figure}

\subsection{Balanced Mixed-Culture Cohort: KDEF/JAFFE}

We next examine a balanced mixed population composed of five Western (KDEF) and five Asian (JAFFE) agents. Figure~\ref{fig:mixed_5_5} illustrates the evolution of Macro-F1 during training and evaluation, separating performance into \emph{intra-cultural}, \emph{cross-cultural}, and \emph{global} averages. During the learning phase ($t < 1000$, $\sigma = 0$), intra-cultural recognition improves rapidly for both groups, following a trajectory similar to the monocultural baselines. In contrast, cross-cultural accuracy remains consistently lower, indicating that agents find it more challenging to recognize expressions from the other cultural group even under optimal visual conditions. By the end of training, intra-cultural Macro-F1 stabilizes near $0.70$-$0.75$, while cross-cultural performance plateaus closer to $0.50$, yielding a persistent gap that reflects the inherent cultural specificity of the learned representations.

During evaluation, increasing blur levels exacerbates these differences. Intra-cultural performance degrades gradually as $\sigma$ increases, but remains substantially higher than cross-cultural recognition across all blur levels. Cross-cultural curves exhibit both lower initial values and steeper declines, indicating reduced robustness when agents attempt to classify expressions from a different cultural background under degraded conditions. The global performance curve, which aggregates both intra- and cross-cultural interactions, lies between the two and shows intermediate robustness. These results highlight that even in balanced mixtures, cultural boundaries remain salient: agents perform well with members of their own group, but systematic errors and reduced confidence emerge in cross-cultural interactions, particularly as perceptual quality worsens.

\begin{figure}[ht]
    \centering
    \includegraphics[width=1\columnwidth]{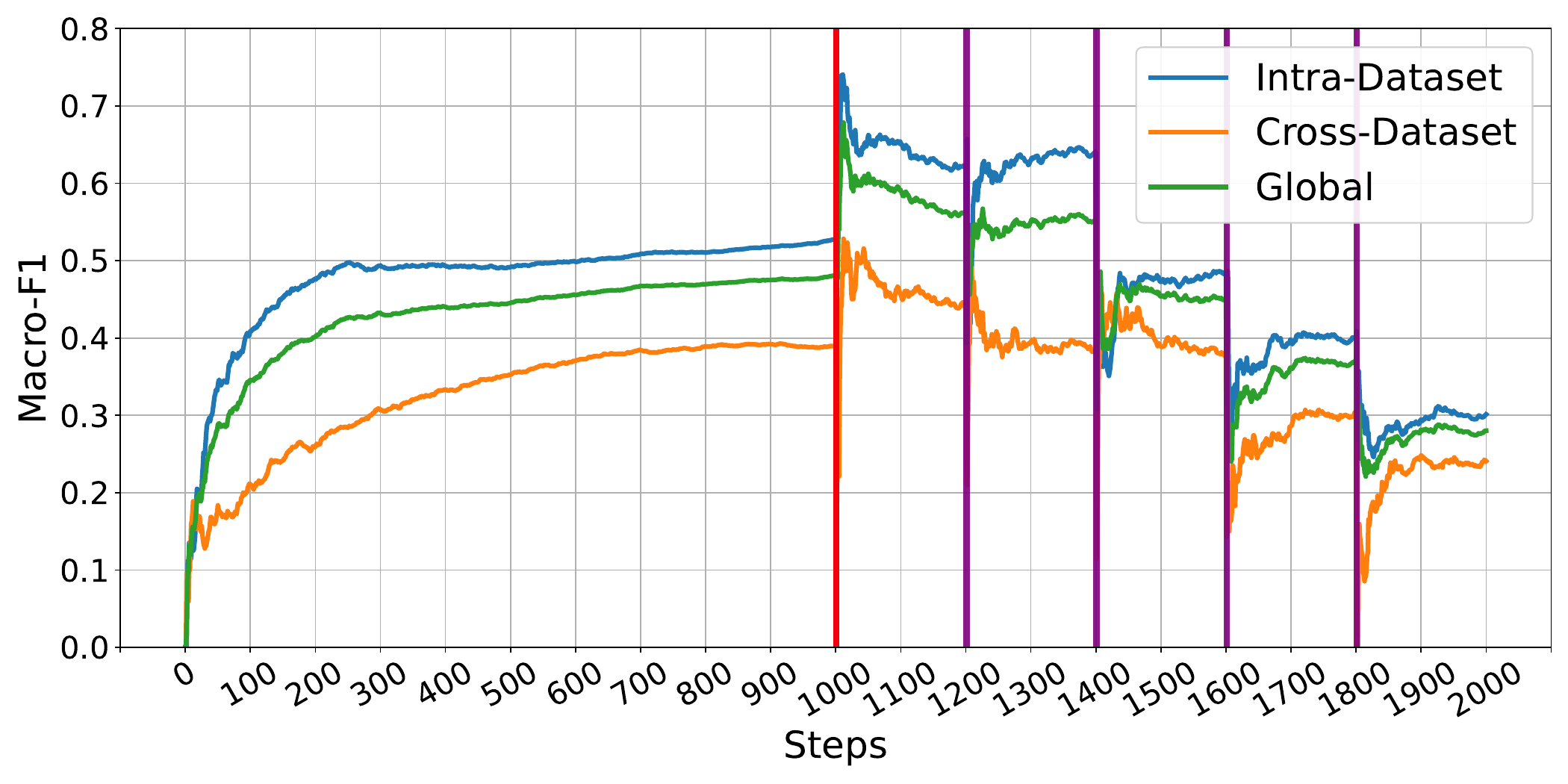}
    \caption{Macro-F1 evolution for an imbalanced mixed population (8 KDEF / 2 JAFFE). The majority group dominates intra-cultural accuracy, while cross-cultural interactions remain weaker, especially under increasing blur.}
    \label{fig:mixed_8_2}
\end{figure}

\subsection{Imbalanced Mixed-Culture Cohort: KDEF-Majority}

We then explore an imbalanced mixed population of eight KDEF (Western) and two JAFFE (Asian) agents. Figure~\ref{fig:mixed_8_2} presents the evolution of intra-cultural, cross-cultural, and global Macro-F1 measures over training and test. During the learning stage ($t < 1000$, $\sigma = 0$), the intra-cultural curve increases rapidly, achieving high accuracy levels comparable to those of the KDEF-only baseline scenario. This is expected, given that interactions between Western agents are monopolized, and thus their classifiers can benefit from rich, culturally homogeneous training material. In contrast, cross-cultural performance remains at substantially lower levels over training, corresponding to the narrow experience of JAFFE agents with Western faces and vice versa. By the end of the learning stage, the gap between intra- and cross-cultural recognition is larger than with a balanced population, with cross-cultural Macro-F1 remaining stuck at around $0.50$ despite high intra-cultural accuracy.

As blur increases during testing, this difference persists and grows. Intra-cultural recognition within the dominant KDEF group remains relatively stable up to medium blur intensities ($\sigma \leq 2$), and performance curves demonstrate the identical gradual degradation as in monocultural KDEF populations. Conversely, cross-cultural accuracy degrades more sharply, especially at high $\sigma$ values, where minority JAFFE agents experience both diminished visual quality and limited training vulnerability. As a result, the global performance curve is significantly influenced by the leading group, remaining relatively high in relation to cross-cultural performance, but masking underlying disadvantages among minority groups. These results indicate that majority dominance boosts intra-cultural dominance: the leading culture sets the performance baseline, and minority–majority cross-cultural interactions consistently introduce error sources, especially under adverse perceptual conditions.

\begin{figure}[ht]
    \centering
    \includegraphics[width=1\columnwidth]{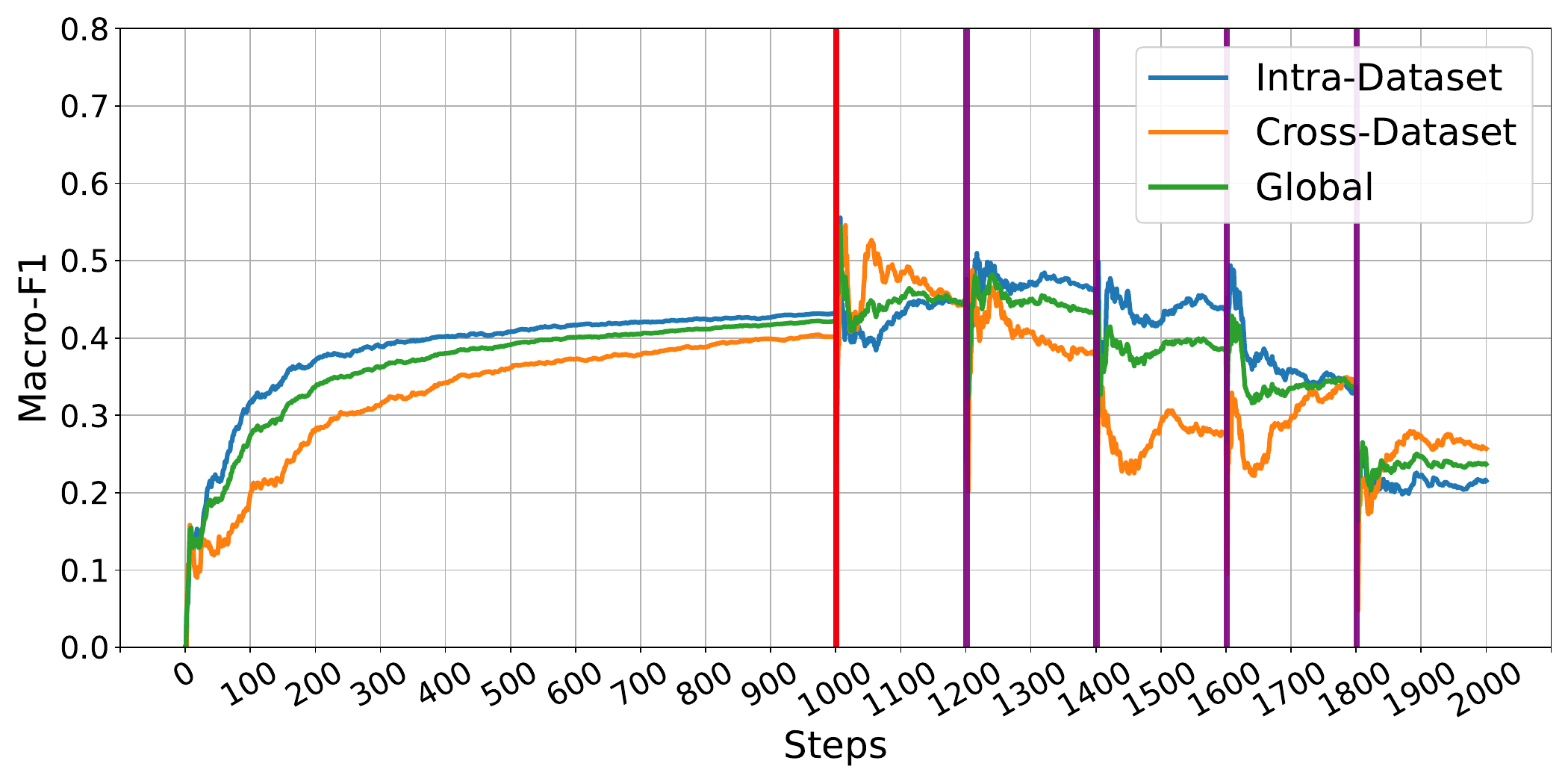}
    \caption{Macro-F1 evolution for an imbalanced mixed population (8 JAFFE / 2 KDEF). JAFFE-majority groups exhibit lower intra-cultural robustness and more pronounced degradation under blur compared to KDEF-majority scenarios.}
    \label{fig:mixed_8_j}
\end{figure}

\subsection{Imbalanced Mixed-Culture Cohort: JAFFE-Majority}

Finally, we examine the complementary imbalanced configuration consisting of eight JAFFE (Asian) agents and two KDEF (Western) agents. Figure~\ref{fig:mixed_8_j} shows the evolution of intra-cultural, cross-cultural, and global Macro-F1 over training and evaluation. During the learning phase ($t < 1000$, $\sigma = 0$), intra-cultural performance increases for the JAFFE-majority population, but reaches a lower asymptote ($\sim 0.58$) compared to the KDEF-majority case. This plateau reflects the intrinsic limitations observed in JAFFE monocultural baselines, where lower image resolution and reduced contrast affect the discriminability of facial expressions. Cross-cultural accuracy remains consistently lower throughout the learning period, stabilizing near $0.50$ due to the small number of KDEF agents and the inherent difficulty of recognizing out-group expressions.

The performance stage points out the symmetry breaking between majority cultures. As blur intensity increases, intra-cultural accuracy for the JAFFE-majority case deteriorates more rapidly than for the KDFC-majority simulation, with noticeable degradation starting at $\sigma = 1$ and a substantial decline for $\sigma \geq 2$. Cross-cultural plots also demonstrate severe degradation due to blur, such that there is little robustness when minority KDFC agents try to recognize JAFFE faces and vice versa. Therefore, the global performance plot remains consistently lower than in the KDFC-majority simulation, which accounts for the lower robustness of JAFFE faces to perceptual degradation per se. These results underscore that the majority culture's identity has a significant influence on system-level performance. When the majority culture itself is inherently weaker (as with JAFFE faces), overall accuracy performance is correspondingly weaker and subject to environmental degradation and cross-cultural interactions, making it especially fragile.

\subsection{Performance Degradation Summary}

\begin{table}[t]
\centering
\caption{Relative performance degradation $\Delta_\sigma$ for each configuration, computed with respect to $\sigma=0$.}
\label{tab:sigma_degradation}
\begin{tabular}{lccccc}
\toprule
\textbf{Condition / Metric} & $\mathbf{\sigma=0}$ & $\mathbf{\sigma=1}$ & $\mathbf{\sigma=2}$ & $\mathbf{\sigma=3}$ & $\mathbf{\sigma=4}$ \\
\midrule
KDEF     &  0.000 & -0.076 &  0.160 &  0.270 &  0.456 \\
JAFFE      &  0.000 &  0.072 &  0.030 &  0.252 &  0.409 \\
\midrule
Balanced (5/5) & 0.000 & 0.014 & 0.246 & 0.332 & 0.411 \\
\midrule
KDEF-majority (2/8) & 0.000 & 0.067 & 0.234 & 0.401 & 0.536 \\
\midrule
JAFFE-majority (8/2) & 0.000 & 0.005 & 0.136 & 0.233 & 0.473 \\
\bottomrule
\end{tabular}
\end{table}

The results in Table~\ref{tab:sigma_degradation} provide a compact summary of the impact of the scheduled perceptual degradation on recognition performance, complementing the dynamic analyses presented in the previous experimental subsections. A clear trend emerges: as blur level ($\sigma$) increases, all configurations experience progressive relative Macro-F1 degradation, although the severity differs depending on the population composition. Monocultural cohorts exhibit distinct degradation profiles: the KDEF population shows a steady decline reaching $\Delta_{\sigma=4}=0.456$, whereas the JAFFE population follows a less monotonic trajectory, with slight increases and decreases at intermediate blur levels, ending at $\Delta_{\sigma=4}=0.409$. This asymmetry is consistent with the temporal curves reported earlier, in which the Japanese population exhibited better performance under low degradation but deteriorated more abruptly at intermediate stages. In contrast, the Western population showed a more uniform downward trend.

Multicultural scenarios reveal intermediate patterns and confirm the interaction between population composition and robustness. In the balanced configuration (5/5), degradation remains moderate up to $\sigma=1$ but accelerates from $\sigma=2$ onward, reaching $\Delta_{\sigma=4}=0.411$, positioning it between the pure KDEF and JAFFE curves. In unbalanced mixtures, the majority group drives the overall trend: the KDEF-majority setting exhibits the steepest degradation at $\sigma=4$ (0.536), suggesting that asymmetric mixing does not mitigate the vulnerabilities of the dominant group under perceptual deterioration but may even amplify them. Conversely, the JAFFE-majority configuration maintains lower degradation levels up to $\sigma=3$ but suffers a sharp increase at $\sigma=4$ (0.473), consistent with the delayed vulnerability observed in the experimental curves. Overall, these patterns confirm that cultural composition and interaction structure significantly shape the robustness of facial expression recognition as visual quality progressively degrades.

\section{Conclusion}

In this study, we introduce an agent-based streaming model for exploring the combined effect of cultural constitution and perceptual decay on facial expression recognition. By placing agents within a lattice space and presenting them with expressions from different cultural databases (KDEF, JAFFE), under a gradually increasing Gaussian blur, we observed dynamic divergent robustness, error models, and calibration between cultural groups. This streaming framework moves beyond fixed train-test partitions, allowing us to capture what happens over time in natural, non-stationary visual environments.

A central component of our approach is the use of CLIP embeddings as a frozen feature extractor, upon which each agent learns a lightweight residual adapter. The reliance on CLIP offers strong generalization and transfer capabilities, but it also introduces implicit inductive biases. Given that CLIP was trained on large-scale web-scraped image–text pairs (e.g., \ WebImageText / LAION pipelines \cite{cherti23}) biased toward Western and English-centric content, it is plausible that its embedding space better represents Western facial appearances and distributions than Asian ones. Indeed, several recent studies have documented representational and demographic biases in CLIP and similar vision–language models, including unequal filtering of non-Western data and skewed associations in the embedding space \cite{hong24}.   Thus, the poorer performance and steeper degradation observed for JAFFE-based agents might be partially attributable to a mismatch between CLIP’s learned embedding biases and the statistical structure of Asian face expression features.

Our experiments reveal consistent trends across cohorts. In monocultural settings, KDEF-only agents showed higher accuracy and blur resistance than JAFFE-only agents, exposing an inherent robustness asymmetry. Mixed populations maintained intra-cultural accuracy but degraded faster in cross-cultural recognition. In imbalanced settings, majority groups dominated performance, while minority interactions degraded most. However, the JAFFE-majority case showed that dominance does not ensure robustness when the embedding bias works against it.

These results underline key methodological and practical points. Methodologically, they highlight that frozen feature extractors can encode structural bias even when adapters are trained fairly. Practically, real-world FER systems must assess group-level resilience under visual degradation and consider mitigation at both representation and adapter levels. Future work should explore balanced embedding models, dynamic debiasing, and domain-aware adaptation to enhance cross-cultural robustness and fairness.

\section*{Acknowledgements}
This work is partially funded by project PID2021-122402OB-C22/MICIU/AEI
/10.13039/501100011033 FEDER, UE and by the ACIISI-Gobierno de Canarias and European FEDER funds under project ULPGC Facilities Net and Grant \mbox{EIS 2021 04}.

\bibliographystyle{IEEEtran}

\end{document}